\documentclass[conference]{IEEEtran}
\IEEEoverridecommandlockouts
\usepackage{cite}
\usepackage{amsmath,amssymb,amsfonts}
\usepackage{algpseudocode}
\usepackage{graphicx}
\usepackage{textcomp}
\usepackage{xcolor}
\def\BibTeX{{\rm B\kern-.05em{\sc i\kern-.025em b}\kern-.08em
    T\kern-.1667em\lower.7ex\hbox{E}\kern-.125emX}}
\begin{document}

\title{Federated Ensemble YOLOv5 – A Better Generalized Object Detection Algorithm\\

}

\author{\IEEEauthorblockN{1\textsuperscript{st} Vinit Hegiste}
\IEEEauthorblockA{\textit{Chair of Machine Tools and Control Systems} \\
\textit{RPTU Kaiserslautern-Landau}\\
Kaiserslautern, Germany \\
vinit.hegiste@rptu.de}
\and
\IEEEauthorblockN{2\textsuperscript{nd} Tatjana Legler}
\IEEEauthorblockA{\textit{Chair of Machine Tools and Control Systems} \\
\textit{RPTU Kaiserslautern-Landau}\\
Kaiserslautern, Germany \\
tatjana.legler@rptu.de}
\and
\IEEEauthorblockN{3\textsuperscript{rd} Martin Ruskowski}
\IEEEauthorblockA{\centerline{Innovative Factory Systems (IFS)} \\
\textit{German Research Center for Artificial Intelligence (DFKI)}\\
Kaiserslautern, Germany \\
martin.ruskowski@dfki.de}
}

\maketitle

\begin{abstract}
Federated learning (FL) has gained significant traction as a privacy-preserving algorithm, but the underlying resemblances  of federated learning algorithms like Federated averaging (FedAvg) or Federated SGD (Fed SGD) to ensemble learning algorithms has not been fully explored. The purpose of this paper is to examine the application of FL to object detection as a method to enhance generalizability, and to compare its performance against a centralized training approach for an object detection algorithm.
Specifically, we investigate the performance of a YOLOv5 model trained using FL across multiple clients and employ a random sampling strategy without replacement, so each client holds a portion of the same dataset used for centralized training.
Our experimental results showcase the superior efficiency of the FL object detector's global model in generating accurate bounding boxes for unseen objects, with the test set being a mixture of objects from two distinct clients not represented in the training dataset. These findings suggest that FL can be viewed from an ensemble algorithm perspective, akin to a synergistic blend of Bagging and Boosting techniques. As a result, FL can be seen not only as a method to enhance privacy, but also as a method to enhance the performance of a machine learning model.
\end{abstract}


\section{Introduction}

In recent years, federated learning (FL) has emerged as a promising privacy-preserving algorithm that allows multiple clients to collaboratively train a global model without sharing their raw data \cite{McMahan.2017}. The concept of FL involves aggregating local model updates from distributed clients to create a more robust and accurate global model. While FL has gained significant traction for its privacy benefits, its underlying resemblance to ensemble learning algorithms such as Bagging or Boosting remains unexplored.
The objective of this paper is to investigate the parallels between federated learning algorithms, such as Federated Averaging (FedAvg) and ensemble learning algorithms, and to explore the application of FL to object detection tasks.
The aggregation of weights to achieve a better global model can be compared with the Bagging algorithm, where multiple clients have data belonging to the same class/label, falling in the same feature space. And the process of sending the global model back to all clients to re-train as the starting weights is similar to the boosting algorithm, where the idea is to train weak classifiers to be able to perform better.

In this paper, we take advantage of the collaborative nature of FL to enhance the overall detection capabilities of the model, therefore creating a better generalized and more stable model: federated ensemble learning YOLOv5  (FedEnsemble YOLOv5).
We examine the performance of the popular YOLOv5 (You Only Look Once version 5) algorithm \cite{Jocher.2020} for object detection when trained using ensemble FL with multiple clients in comparison to the traditional centralized training approach.
To accomplish this, we propose a methodology where multiple clients participate in the federated training process, with each client possessing a subset of the same dataset used in centralized training. The data distribution among clients is achieved through random sampling without replacement. This approach allows us to assess the effectiveness of FL for object detection using the YOLOv5 model in an ensemble setting.
By leveraging the collective knowledge of diverse clients, we aim to demonstrate the potential of FL as an ensemble algorithm, exhibiting characteristics similar to Bagging and Boosting techniques.

\section{Related work}
Ensemble learning methods use different approaches to combine the outputs of many (often simpler) models into a single collective output that achieves better performance than any individual model alone \cite{Zhou.2012}. Although the foundation for some of these methods can be traced back several decades \cite{Drucker.1994}  \cite{Schapire.1990}, they are gaining renewed significance due to the importance of machine learning \cite{Dong.2020}. The combination rule depends on the type of problem at hand, the model being used and may range from simple to complex: Calculating the mean value of every output (bagging) or taking the number of votes for a prediction class (voting) are the most intuitive methods, often resulting in already good results \cite{Wan.2013}. There are different types of boosting approaches, all of which aim to reinforce a feature of the outputs between the classifiers, such as giving more weight to previously misclassified samples \cite{Svetnik.2005}.
Stacking is usually a two-step process, that uses multiple base-level models to generate predictions, which are then used as input for a higher-level model to produce the final prediction 
\cite{Wolpert.1992}.

Combining ensemble learning and federated learning can leverage the advantages of both methods, enabling efficient learning from decentralized datasets while benefiting from the predictive power of multiple learners. 
There exist numerous approaches to combining FL and ensemble learning, each with different objectives:
FedBoost introduces boosting to FL for communication cost reduction, while assuming that pre-trained base predictors exist \cite{Hamer.2020}.
\cite{Lin.2020} utilizes knowledge distillation known from continual learning and further developed this into ensemble distillation of the clients, but requires labeled data to be available to the server. 
\cite{Yu.2023} focuses on the improvement of performance of a non-IID image classification by following a stacking approach. 
For object detection in our research, we adopted the YOLOv5 algorithm \cite{Jocher.2020}, which is an improved implementation of the YOLOv3 algorithm \cite{Redmon.2016} using the PyTorch framework. The selection of YOLOv5 as our object detection algorithm aligns with the methodology employed in a previous study \cite{Hegiste.2023}, allowing for direct comparisons between the outcomes presented in both papers. This choice ensures consistency and facilitates a comprehensive analysis of the results obtained in our current research in relation to the earlier work.
Several others have also developed more advanced concepts, however, none have explored the fundamental analogies between ensemble and federated learning.

\section{Methodology} \label{method}
In this section, we elaborate on our approach to federated ensemble learning by structuring clients that operate as weak learners. We segment the principal dataset into a number of mutually exclusive subsets, each corresponding to a specific client.
This division of data, sets the stage for implementing FedAvg algorithm, which aims to empower these weak learners by enabling them to acquire distinctive features and characteristics by utilizing the global models. Interestingly, each round of communication serves to effectively boost the performance of these learners, echoing the mechanics of Boosting algorithms. This systematic approach culminates in the creation of a robust global model that surpasses its counterparts in terms of performance, displaying superior generalization capabilities in classifying and drawing bounding boxes over the target objects.

Inspired by the ensemble nature of federated learning, we designed an experiment revolving around the goal of achieving an effective object detection model. Our model leverages the YOLOv5 algorithm, utilizing a dataset identical to the one employed in centralized YOLOv5 training. In contrast to conventional FL, the centralized dataset is randomly divided into partitions equivalent to the number of clients. Each client then engages in training a federated global model, where active participation in every communication round must be ensured. This approach allows us to make the most of the inherent ensemble behavior of federated learning, crafting an innovative solution for object detection that builds upon the strengths of existing methodologies.

The centralized dataset is divided into three subsets for training, validation, and testing purposes.
Specifically, 80\% of the total dataset is allocated as the training dataset, while 10\% each is reserved for validation and testing. When distributing the centralized dataset to multiple clients, the same images that were part of the training subset used for training the centralized YOLOv5 model are provided.
To simplify the process, we consider a fixed number of three clients and provide them with identical validation and test datasets that were originally utilized during the centralized training.

This distribution strategy remains consistent for both the dataset discussed in Section \ref{dataset}.
In our preliminary experiments, we conducted a test using the federated ensemble (FedEnsemble) algorithm on a dataset comprising USB-Sticks, which was previously presented in \cite{Hegiste.2022}. The results indicated that the FedEnsemble technique yielded superior accuracy compared to a centrally trained USB quality classification model.
Encouraged by this outcome, we extended our work to include federated ensemble object detection, aiming to address practical applications beyond image classification.

\begin{figure*}[h]
    \centering
    \includegraphics[width= \textwidth]{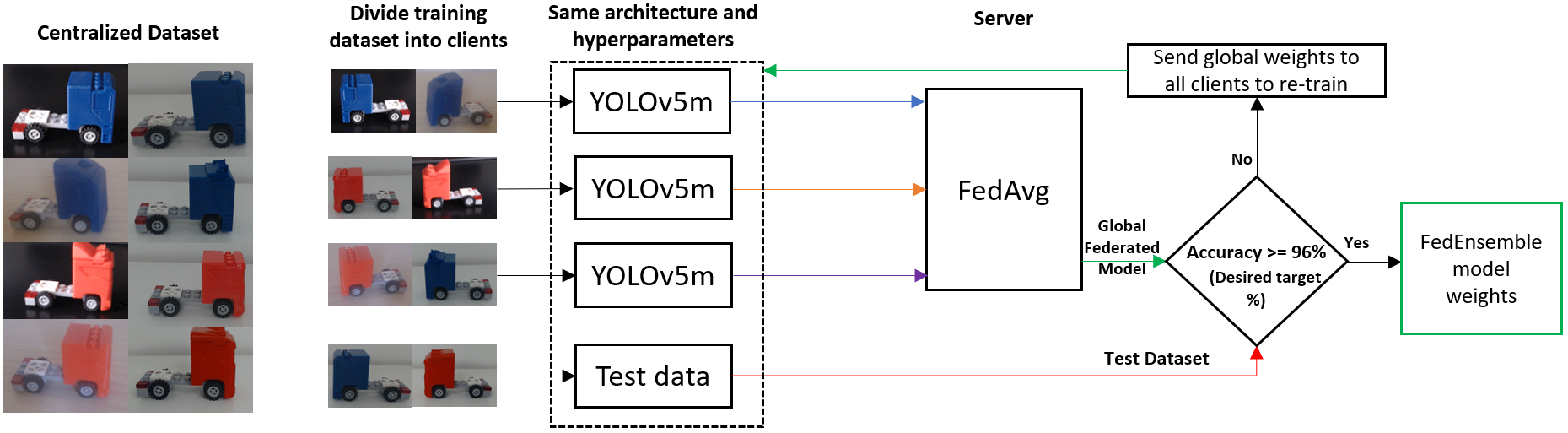}
    \caption{FedEnsemble algorithm with example of cabin dataset}
    \label{fig:fedensemble_flowchart}
\end{figure*}



\subsection{Ensemble Algorithm}
In order to leverage the ensemble behavior of federated learning, we conducted experiments with a new algorithm to develop a federated ensemble object detection model based on YOLOv5. This algorithm utilizes the same dataset that was employed for the centralized training of YOLOv5. The dataset is divided into 3 clients and the distribution strategy is the same as mentioned in Section \ref{method}. The number of clients can be increased based on the size of the centralized training dataset.

The algorithm follows the steps outlined below:
\begin{algorithmic}[1]
\State Shuffle the centralized data with YOLO annotations and divide it into 'n' mutually exclusive datasets, where 'n' represents the number of clients. The validation and test datasets remain unchanged.
\State Utilize the same YOLOv5 model architecture (e.g., YOLOv5m or YOLOv5l) and consistent hyperparameters (e.g., optimizer, batch size, learning rate, local epochs) for all clients.
\State Train each client individually using their respective local dataset for a specified number of epochs and save the weights of the last epoch.
\State Send the weights of all clients to the server and perform federated averaging to create a global model. It is important to note that all clients actively participate in each communication round.
\State Evaluate the accuracy of the global model on the test dataset.
\State Terminate the process when the global model achieves the desired accuracy on the test dataset.
\end{algorithmic}

Figure \ref{fig:fedensemble_flowchart} depicts the process flow diagram of the above-mentioned algorithm.
The resulting global model is then compared to the normally trained YOLOv5 model, where identical hyperparameters are maintained. The outcomes of this comparison can be found in Section \ref{results}. 
To expedite the process, clients can be executed on parallel GPUs while running the server code on the same system. This configuration can significantly enhance the computational efficiency of the federated ensemble algorithm. Since the federated ensemble learning approach does not tackle privacy issues as the subset used for training by every client is a part of a centralized dataset; hence this parallelization strategy can be employed without any concerns in that regard.

\subsection{Datasets} \label{dataset}

\begin{figure}[h]
    \centering
    \includegraphics[width= 0.48\textwidth]{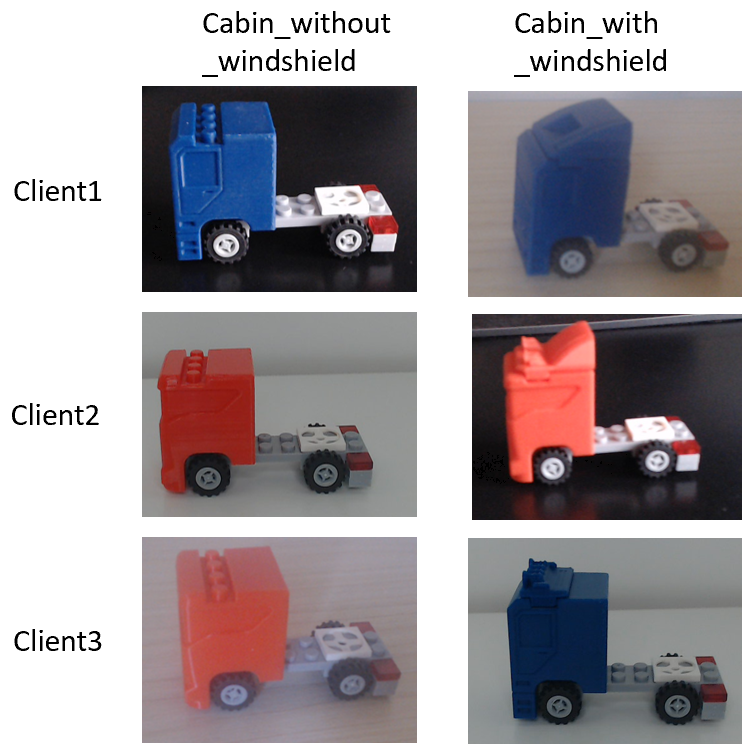}
    \caption{A small example of distribution of centralized cabin dataset into three clients}
    \label{fig:cabin_dataset}
\end{figure}

In this paper, we experiment with two different custom datasets, which belong to our manufacturing setting, to analyze the robustness of the federated ensemble algorithm on multiple use-cases and scenarios. The first preliminary tests were done on the dataset of USB-Sticks, previously presented in \cite{Hegiste.2022} and \cite{Hegiste.2023}.
The first dataset employed in our study is the truck cabin dataset, which encompasses two distinct design variations of miniature truck cabins in red and blue colors. 
Figure \ref{fig:cabin_dataset} illustrates this dataset, which is identical to the one utilized in a previous study \cite{Hegiste.2023}. The centralized cabin dataset comprises a total of 1200 images, with 600 images categorized as 'Cabin\_without\_windshield' and the remaining 600 images labeled as 'Cabin\_with\_windshield'. 
The dataset comprises four different windshield designs. The training data encompasses blue cabins with blue windshields of type A and B, along with red cabins featuring red windshields of type C and D.
Following the distribution strategy outlined in the methodology section, this dataset is divided among three clients by shuffling the training subset originally used for training the centralized YOLOv5 model. Consequently, the clients do not possess an equal number of instances for each class, as they are randomly assigned.

\begin{figure}[h]
    \centering
    \includegraphics[width= 0.48\textwidth]{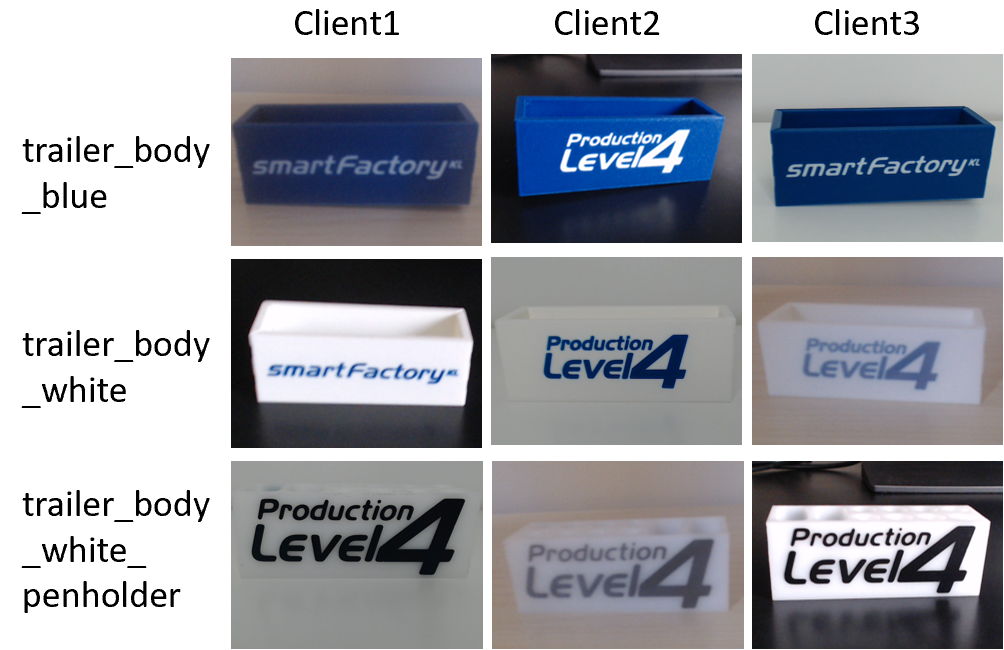}
    \caption{A small example of distribution of centralized trailer dataset into three clients}
    \label{fig:trailer_dataset}
\end{figure}

\begin{table*}[h]
\centering
\caption{Training dataset distribution of trailer dataset (No. of images)}
\begin{tabular}{|c|c|c|c|c|}
\hline
Dataset & Training images & White trailer & Blue trailer & Penholder trailer \\ \hline
Centralized & 720 & 246 & 233 & 241 \\ \hline
Client1 & 241 & 78 & 79 & 84 \\ \hline
Client2 & 240 & 95 & 60 & 85 \\ \hline
Client3 & 239 & 73 & 94 & 72 \\ \hline
\end{tabular}
\label{table1}
\end{table*}

The trailer dataset used in this study comprises a total of 900 images. Each class in the dataset consists of 300 images, and there are three different backgrounds included. Additionally, the dataset includes some deliberately introduced blurry images to enhance the robustness of the trained model. Figure \ref{fig:trailer_dataset} showcases a small subset of the dataset, providing a glimpse into its content.
To facilitate the application of the federated ensemble algorithm, we partitioned the training dataset of the centralized trailer dataset into three disjoint subsets. This partitioning ensures that the validation and test datasets remain consistent across all experiments, enabling fair and accurate comparisons. It is important to note that during the capture of the trailer dataset, the images were obtained without including the chassis. 
Therefore, when evaluating the model's performance on trailers with chassis, interesting and informative results are observed.
For the purpose of data distribution analysis, the centralized training dataset comprises 80\% of the total dataset, resulting in a selection of 720 images. The distribution of these images within each class is randomized. For a comprehensive understanding of the dataset's detailed distribution, refer to Table \ref{table1}. 

\section{Implementation and Experiments} \label{imp and exp}
In this section, we outline the implementation details of the FedEnsemble algorithm for training the federated YOLOv5 models. As mentioned in Section \ref{method}, we distribute a centralized dataset for YOLOv5 into multiple clients (in our case, three clients). Each client utilizes the YOLOv5m architecture with the same set of hyperparameters.
The training process begins by training each client's model for 10 epochs on its local dataset. After completing the local training, the client's weights are sent to a central server, where the FedAvg algorithm is applied to obtain the global weights. However, before sending the weights back to the clients, the global federated model is evaluated using the test dataset. This evaluation allows us to monitor the progress and determine if the model has achieved the desired metrics.
If the global model's performance meets the desired target, the FedEnsemble learning process is stopped. However, if the metrics do not meet the target, the updated weights are sent back to each client, serving as the starting point for the subsequent round of training. It is important to note that in FedEnsemble learning, each client participates in every communication round.
To illustrate this process in a more visual manner, Figure \ref{fig:fedensemble_flowchart} presents a flowchart depicting the steps involved in the FedEnsemble algorithm.
For the cabin dataset, the centralized model was trained for 150 epochs, while for the trailer dataset, it was trained for 100 epochs (after several experimentation this epoch number was selected based on the best performing model). The FedEnsemble model for the cabin dataset yielded good results when trained with 15 local epochs and 5 communication rounds. On the other hand, the best results for the trailer dataset were achieved with 10 local epochs and 4 communication rounds. Once again, this local epochs and communication rounds number was a result of various experimentation. Also in Figure \ref{fig:fedensemble_flowchart}, the accuracy 96\% was set due to the accuracy of YOLOv5 model achieved from normal training. These parameters can vary depending upon the requirement or custom use case.

In the cabin dataset, both the centralized trained YOLOv5 model and the FedEnsemble model exhibit excellent performance on the validation and test datasets. 
To facilitate comparison, we constructed test combinations consisting of cabins and windshields not present in the training dataset, such as blue cabins with blue windshields of type C and D, and red cabins with red windshields of type A and B (as previously utilized in \cite{Hegiste.2023}).
Both models demonstrate high accuracy and accurately predict bounding boxes on this test dataset.
However, it is important to note that the test dataset was generated using a similar background and environmental setting as the training dataset, aiming to assess the models' robustness in accurately classifying objects and drawing precise bounding boxes.
It is worth mentioning two key distinctions between the truck cabin and trailer datasets. Firstly, all cabin images in the truck cabin dataset feature cabins on top of a chassis, whereas the trailer dataset includes chassis images as well. Secondly, both datasets solely contain individual objects, and neither dataset includes fully assembled trucks comprising both cabins and trailers. Based on these considerations, we designed the following experiments for this study:
\begin{enumerate}
    \item Conduct a comparison between the centralized trained YOLOv5 and FedEnsemble Cabin models using a mix of cabin combinations (with and without chassis) for live classification.
    \item Evaluate the outputs of the centralized trained YOLOv5 and FedEnsemble Cabin models on cabins assembled with various trailer types (fully assembled trucks).
    \item Analyze the outputs of the centralized trained YOLOv5 and FedEnsemble Cabin models using images from the quality inspection module on the demonstrator.
    \item Compare the performance of the centralized trained YOLOv5 and FedEnsemble Trailer models for trailers (with and without chassis) in live classification.
    \item Examine the outputs of the centralized trained YOLOv5 and FedEnsemble Trailer models on cabins assembled with different trailer types (fully assembled trucks).
    \item Analyze the outputs of the centralized trained YOLOv5 and FedEnsemble Trailer models using images from the quality inspection module on the demonstrator.
\end{enumerate}
The quality inspection module is an integral part of the demonstrator located at SmartFactory-Kaiserslautern (SF-KL), which showcases the future of manufacturing with production level 4 capabilities.

\section{Results} \label{results}

\begin{figure}[h]
    \centering
    \includegraphics[width= 0.48\textwidth]{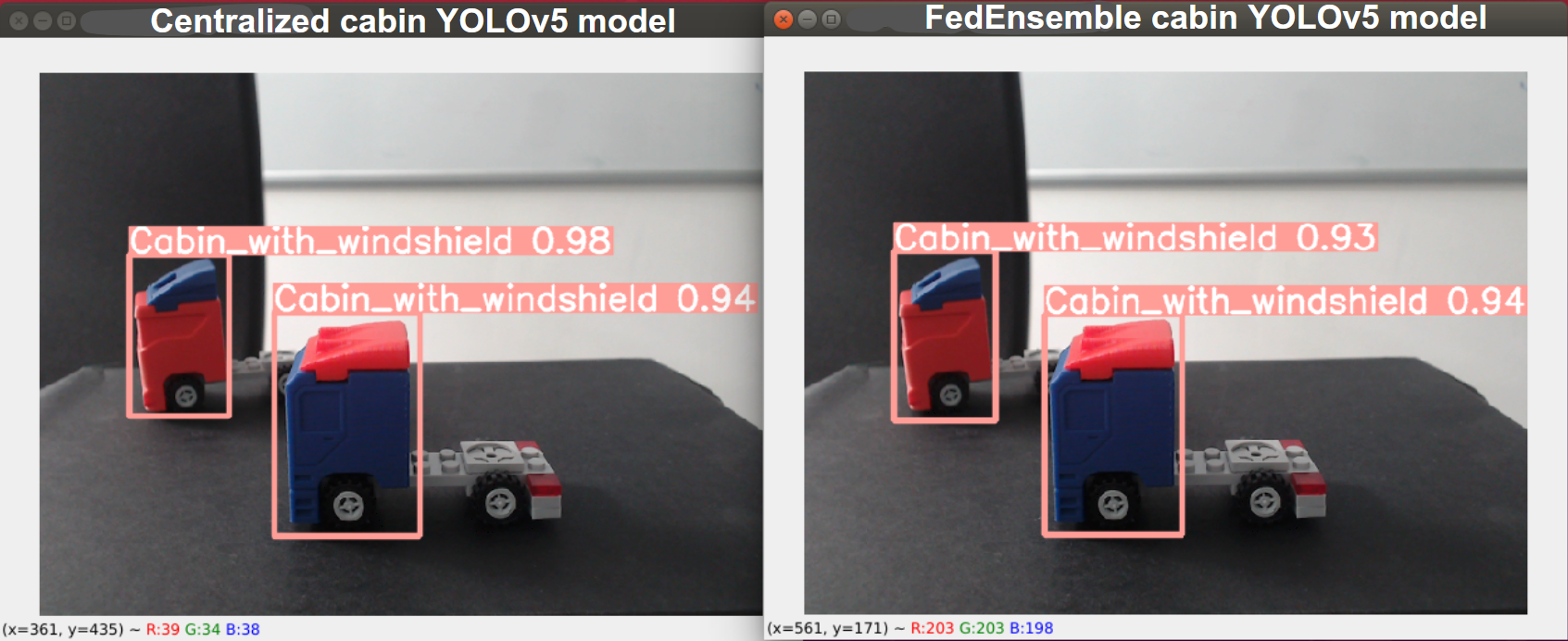}
    \caption{Comparison of models trained using normal YOLOv5 (left) vs Federated Ensemble YOLOv5 (right)when object has a similar orientation, but a different combination not present in the dataset (Cabin with different color type windshield with chassis)}
    \label{fig:experiment1}
\end{figure}

\begin{figure}[h]
    \centering
    \includegraphics[width= 0.48\textwidth]{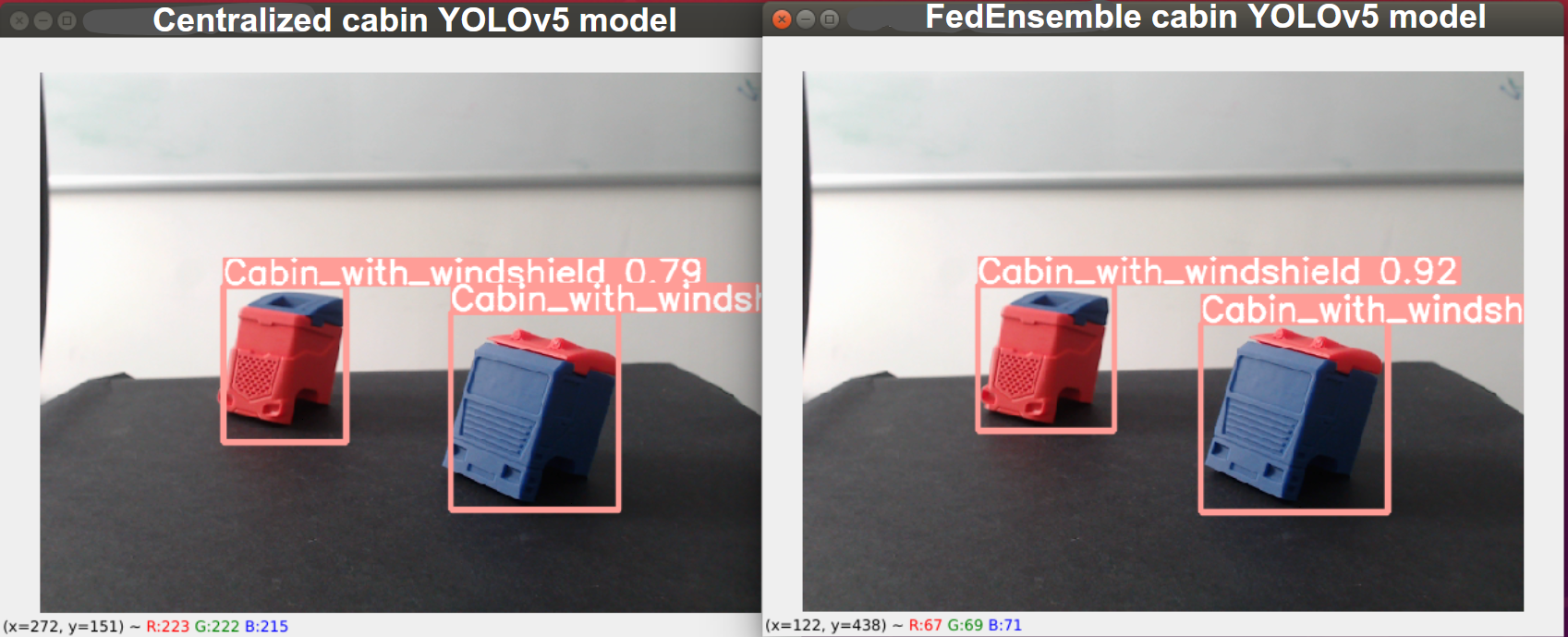}
    \caption{Comparison of models trained using normal YOLOv5 (left) vs FedEnsemble YOLOv5 (right) when object has a different combination not present in the dataset (Cabin with different color type windshield without chassis) }
    \label{fig:experiment1_2}
\end{figure}

In this section, we present the results of the experiments conducted in this paper. 
To facilitate better visualization, the model outputs are depicted with different color bounding boxes (BB). For the cabin models, the 'red color' BB represents the 'Cabin\_without\_windshield' class, while the 'pink color' BB represents the 'Cabin\_with\_windshield' class. Similarly, for the trailer model, the 'red color' BB corresponds to the 'trailer\_body\_blue' class, the 'pink color' BB corresponds to the 'trailer\_body\_white' class, and the 'orange color' BB corresponds to the 'trailer\_body\_white\_penholder' class.
For experiments number 2 and 5, the test images remain consistent for the cabin and trailer combinations to enable better comparison \footnote{The output images have been cropped and enlarged for improved visibility}.
Figures \ref{fig:experiment1} and \ref{fig:experiment1_2} present the outcomes of experiment number 1. Both figures consist of two windows, wherein the left window displays the output from the centrally trained YOLOv5 model, while the right window showcases the output from the FedEnsemble model on a live video frame.
In Figure \ref{fig:experiment1}, the frames feature two cabins with a mixed combination (including a chassis), illustrating the effective classification and accurate bounding box generation by both models.
Moving on to Figure \ref{fig:experiment1_2}, the results are noteworthy as the cabins lack a chassis, resulting in object orientations that differ from those in the training dataset.
While the centrally trained YOLOv5 model correctly classifies the objects, it produces imprecise bounding boxes that lead to partial cropping of the objects. In contrast, the FedEnsemble model exhibits exceptional performance by accurately classifying the objects with high confidence scores and generating precise bounding boxes around both objects in the frame.


\begin{figure}[h]
    \centering
    \includegraphics[width= 0.48\textwidth]{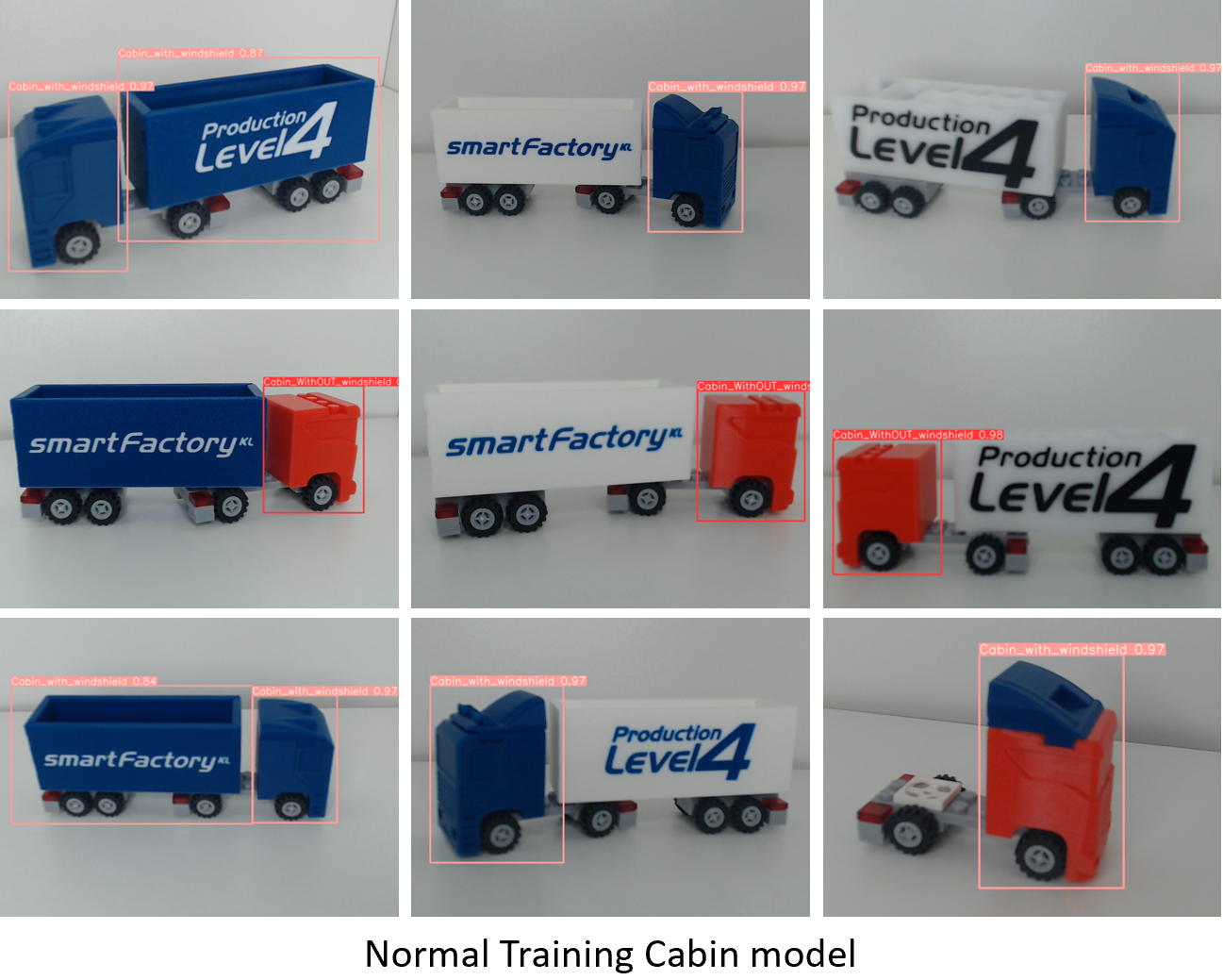}
    \caption{Output of cabin model trained with Normal YOLOv5 algorithm on cabins assembled with different trailer types}
    \label{fig:experiment2}
\end{figure}

\begin{figure}[h]
    \centering
    \includegraphics[width= 0.48\textwidth]{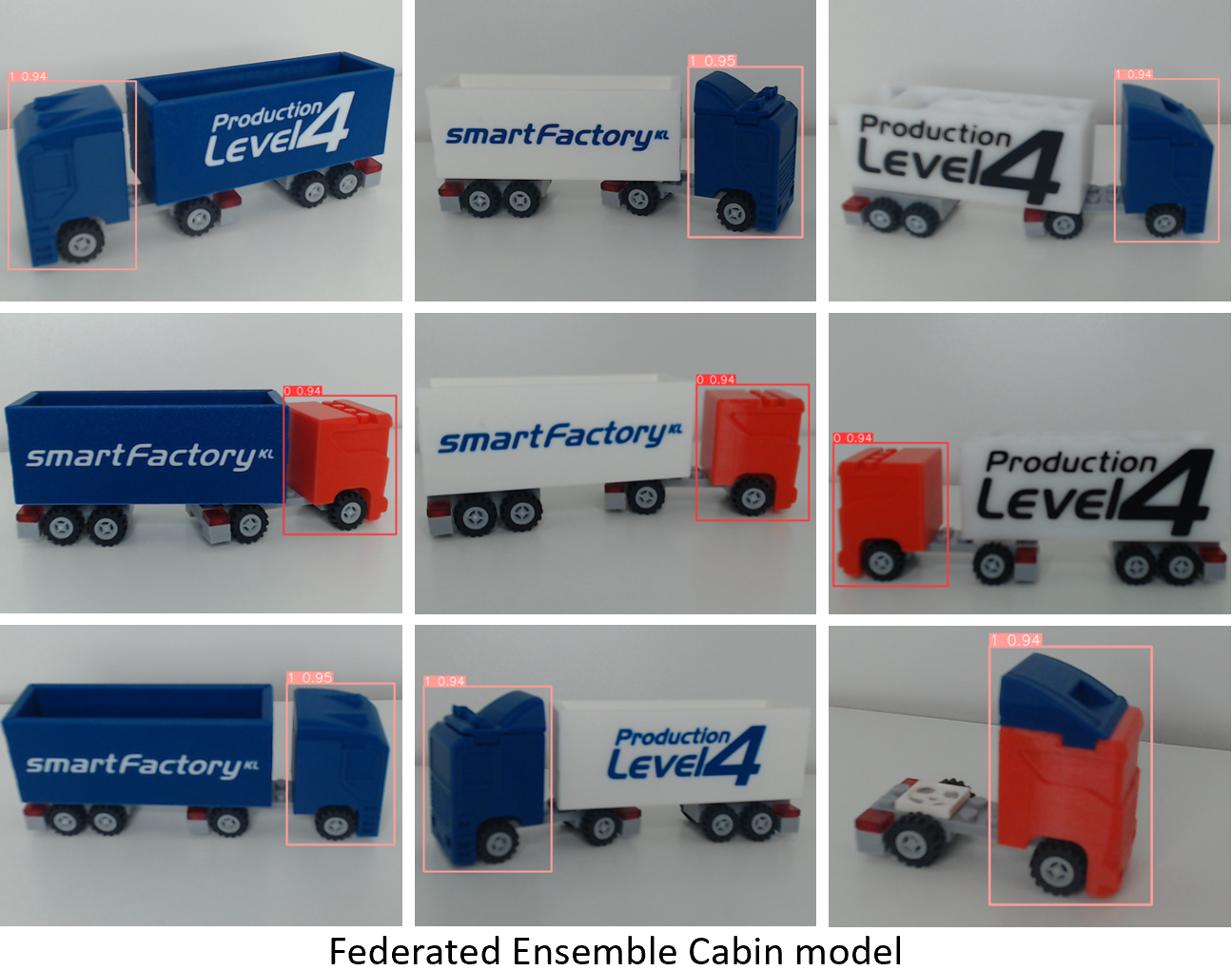}
    \caption{Output of cabin model trained with Federated Ensemble YOLOv5 algorithm on cabins assembled with different trailer types}
    \label{fig:experiment2_2}
\end{figure}

For experiment number 2, both cabins were assembled with different trailer types, and the results of this experiment can be seen in figure \ref{fig:experiment2} and \ref{fig:experiment2_2}. Figure \ref{fig:experiment2} shows output of YOLOv5 cabin model on various cabin and trailer combinations.
The model produces good results apart from the images with blue trailers. The model draws a BB over the blue trailer body and classifies it as 'Cabin\_with\_windshield', leading to false positives.
Figure \ref{fig:experiment2_2} shows the output of FedEnsemble model on the same test images, and the model is able to predict precise BB without producing any false positives on trailer objects.


\begin{figure}[h]
    \centering
    \includegraphics[width= 0.48\textwidth]{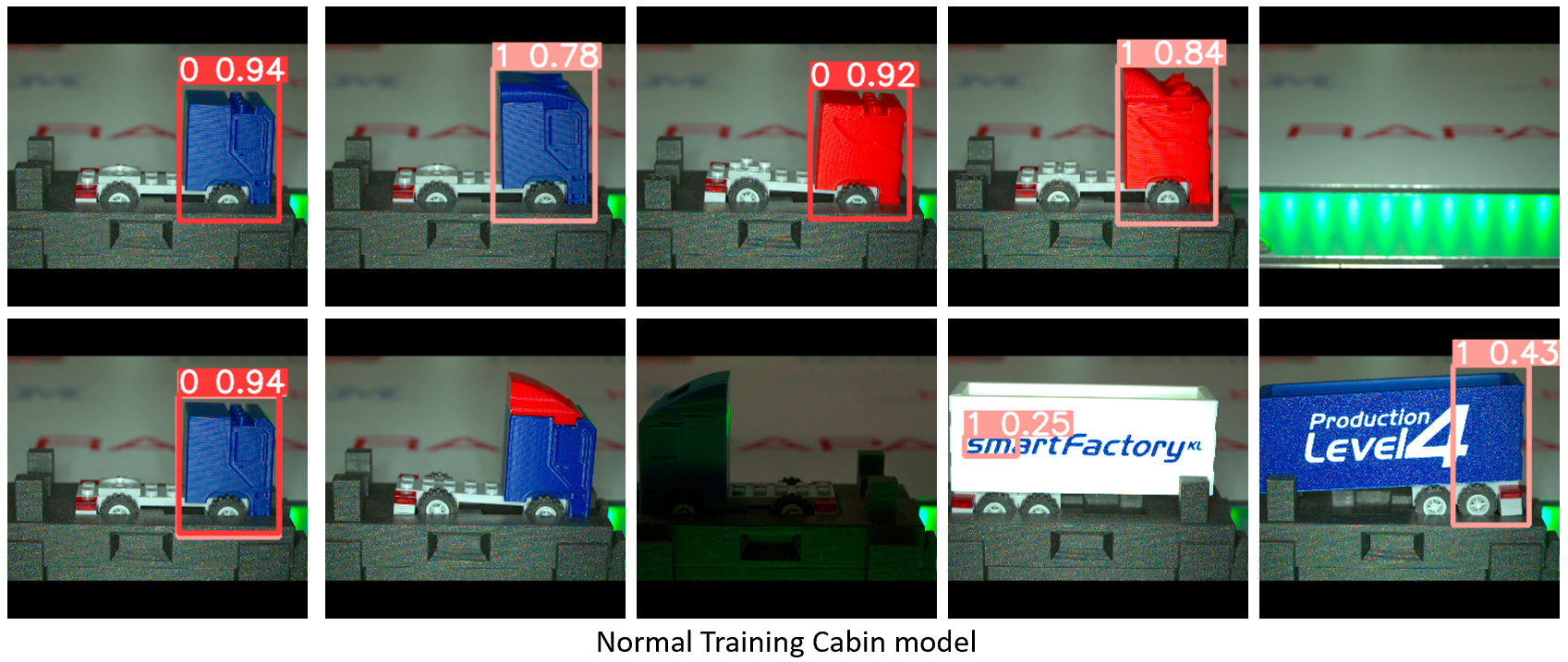}
    \caption{Output of model trained with Normal YOLOv5 algorithm on images from the Demonstrator (0: Cabin\_without\_windshield, 1: Cabin\_with\_windshield)}
    \label{fig:experiment3}
\end{figure}

The results obtained from Experiment 3 are particularly intriguing due to the significant differences in image environment, background, and lighting conditions compared to the images present in the training dataset. Figure \ref{fig:experiment3} displays the output of the YOLOv5 cabin model on test images captured from the demonstrator. While the model performs well on cabin combinations similar to those in the training dataset, it struggles to detect the combination of blue cabin with red windshield and also produces false positives on trailer objects.
In contrast, the results of the FedEnsemble cabin model demonstrate significant improvements. The bounding boxes predicted by the model exhibit high accuracy, with no false positives on trailer objects. Notably, one test image showcases a remarkable outcome: despite the flashlight of the quality inspection module being off and the object being barely visible to the human eye at first glance, the FedEnsemble cabin model is able to correctly classify the object and draw a precise bounding box around it.

\begin{figure}[h]
    \centering
    \includegraphics[width= 0.48\textwidth]{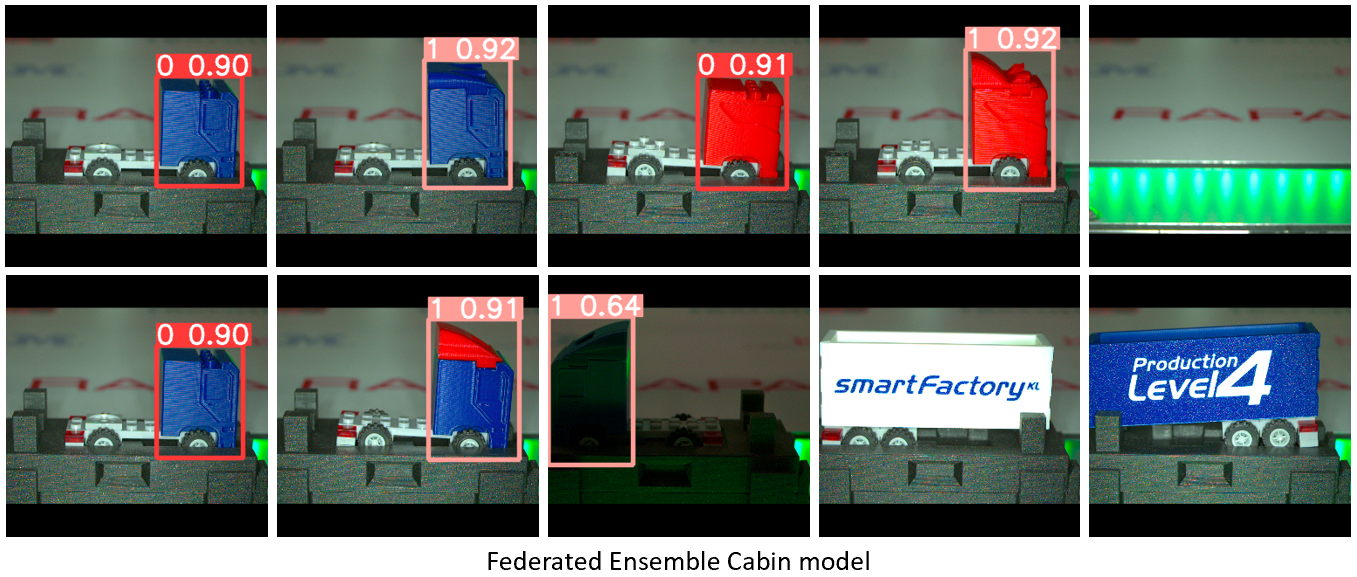}
    \caption{Output of model trained with Federated Ensemble YOLOv5 algorithm on images from the Demonstrator (0: Cabin\_without\_windshield, 1: Cabin\_with\_windshield)}
    \label{fig:experiment3_2}
\end{figure}

\begin{figure}[h]
    \centering
    \includegraphics[width= 0.48\textwidth]{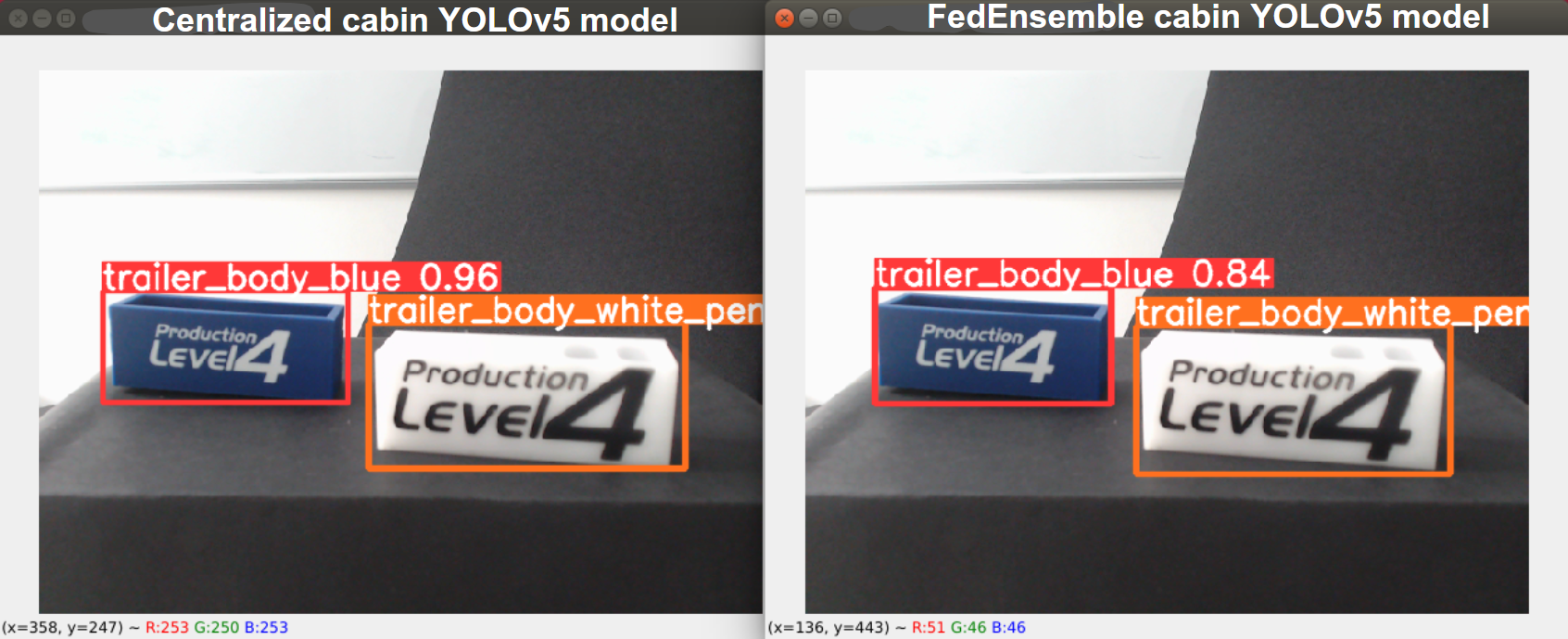}
    \caption{Comparison of models trained using normal YOLOv5 (left) Federated Ensemble YOLOv5 (right)  when the object (trailer) orientation is similar to the training dataset}
    \label{fig:experiment4}
\end{figure}

\begin{figure}[h]
    \centering
    \includegraphics[width= 0.4\textwidth]{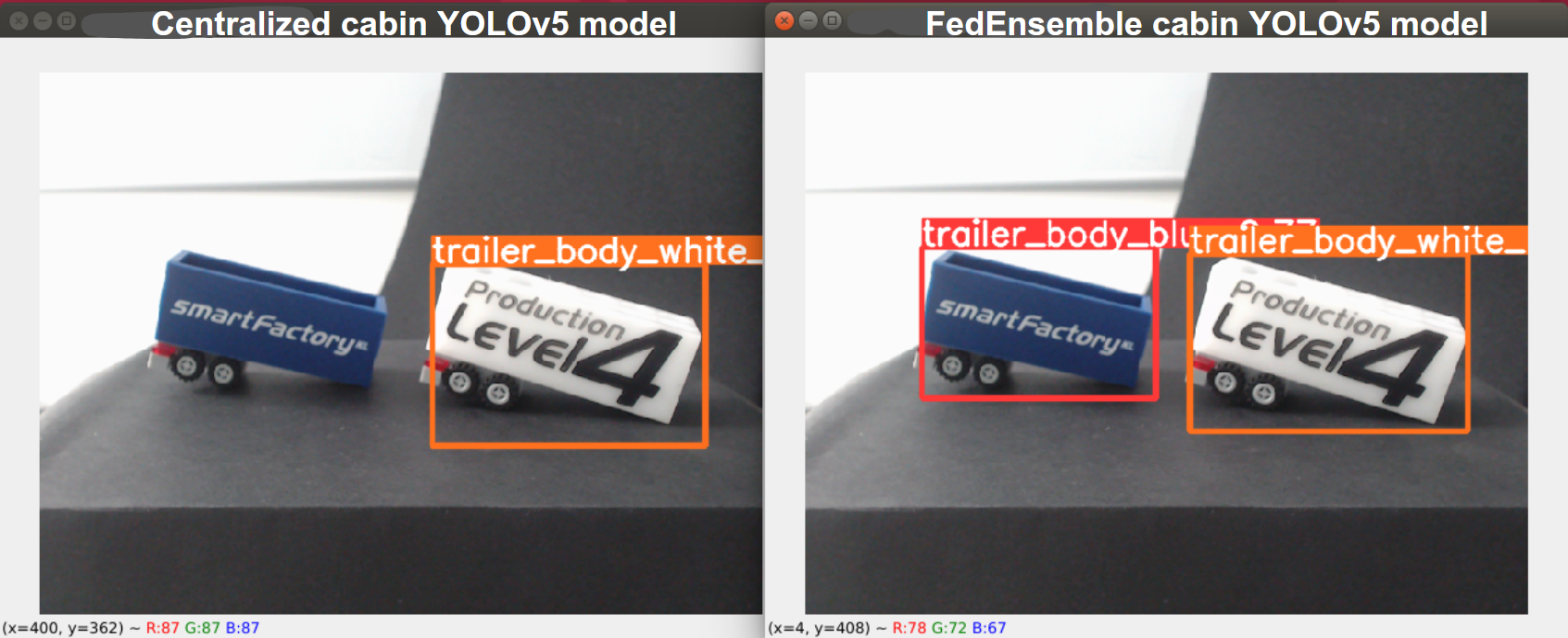}
    \caption{Comparison of models trained using normal YOLOv5 (left) Federated Ensemble YOLOv5 (right) when the object (trailer attached to a chassis) orientation is not similar to the training images}
    \label{fig:experiment4_2}
\end{figure}

Experiments 4, 5, and 6 focus on the trailer dataset models, with an emphasis on maintaining consistency in the test dataset compared to the previous results. Figure \ref{fig:experiment4} and \ref{fig:experiment4_2} illustrate the output of the YOLOv5 and FedEnsemble trailer models, respectively.
Both figures exhibit different orientations of trailer bodies (with and without chassis) in the context of live classification.
It is observed that both models correctly classify and predict trailer objects without a chassis (image orientation similar to the training dataset).
However, as depicted in Figure \ref{fig:experiment4_2}, the YOLOv5 trailer model struggles to classify the blue trailer accurately and crops out a portion of the penholder trailer when predicting the bounding box. In contrast, the FedEnsemble model demonstrates superior performance by drawing precise bounding boxes around both objects in the frame, further highlighting its superiority over the normal centralized trained YOLOv5 model.


\begin{figure}[h]
    \centering
    \includegraphics[width= 0.48\textwidth]{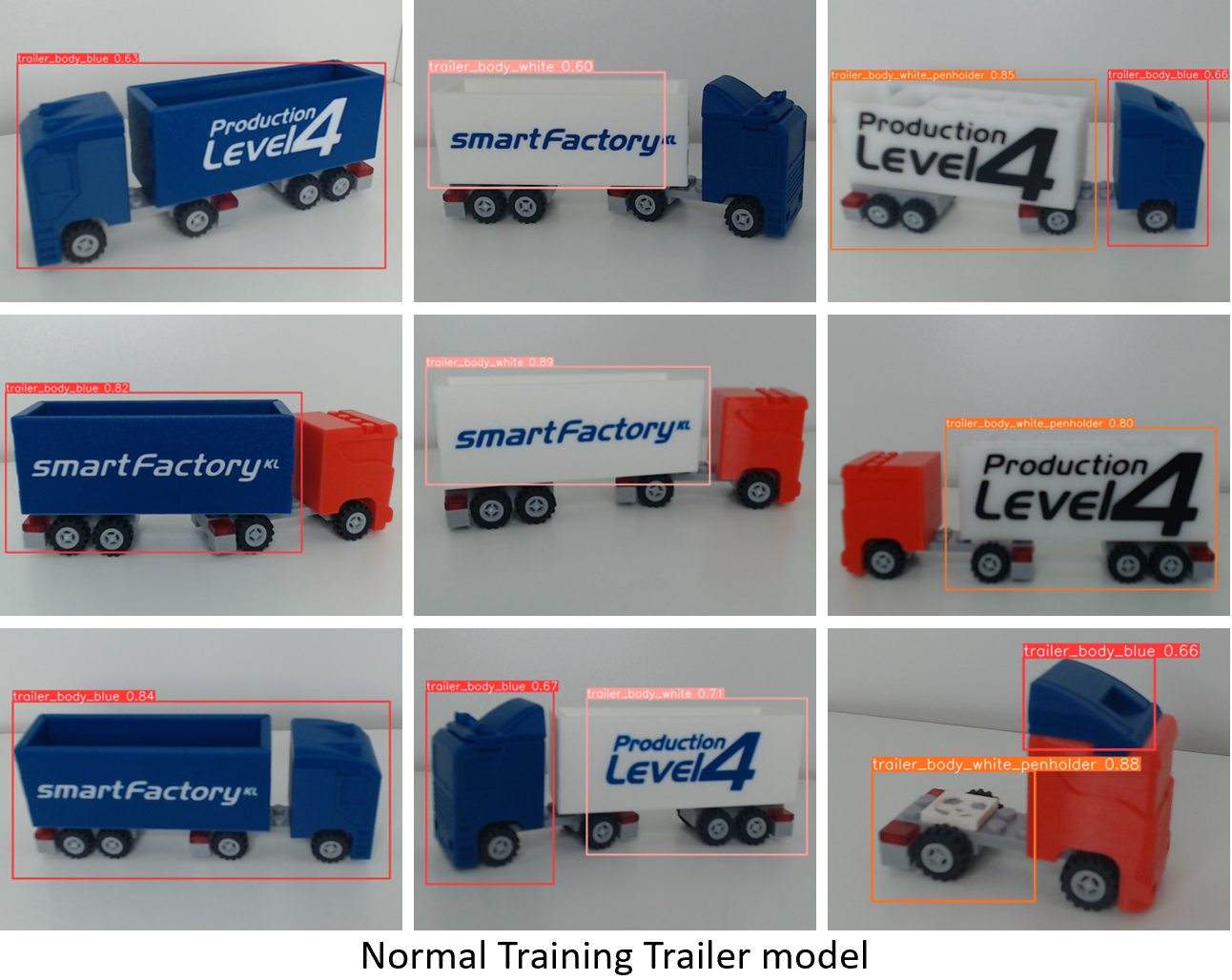}
    \caption{Output of trailer model trained with Normal YOLOv5 algorithm on cabins assembled with different trailer types}
    \label{fig:experiment5}
\end{figure}

\begin{figure}[h]
    \centering
    \includegraphics[width= 0.48\textwidth]{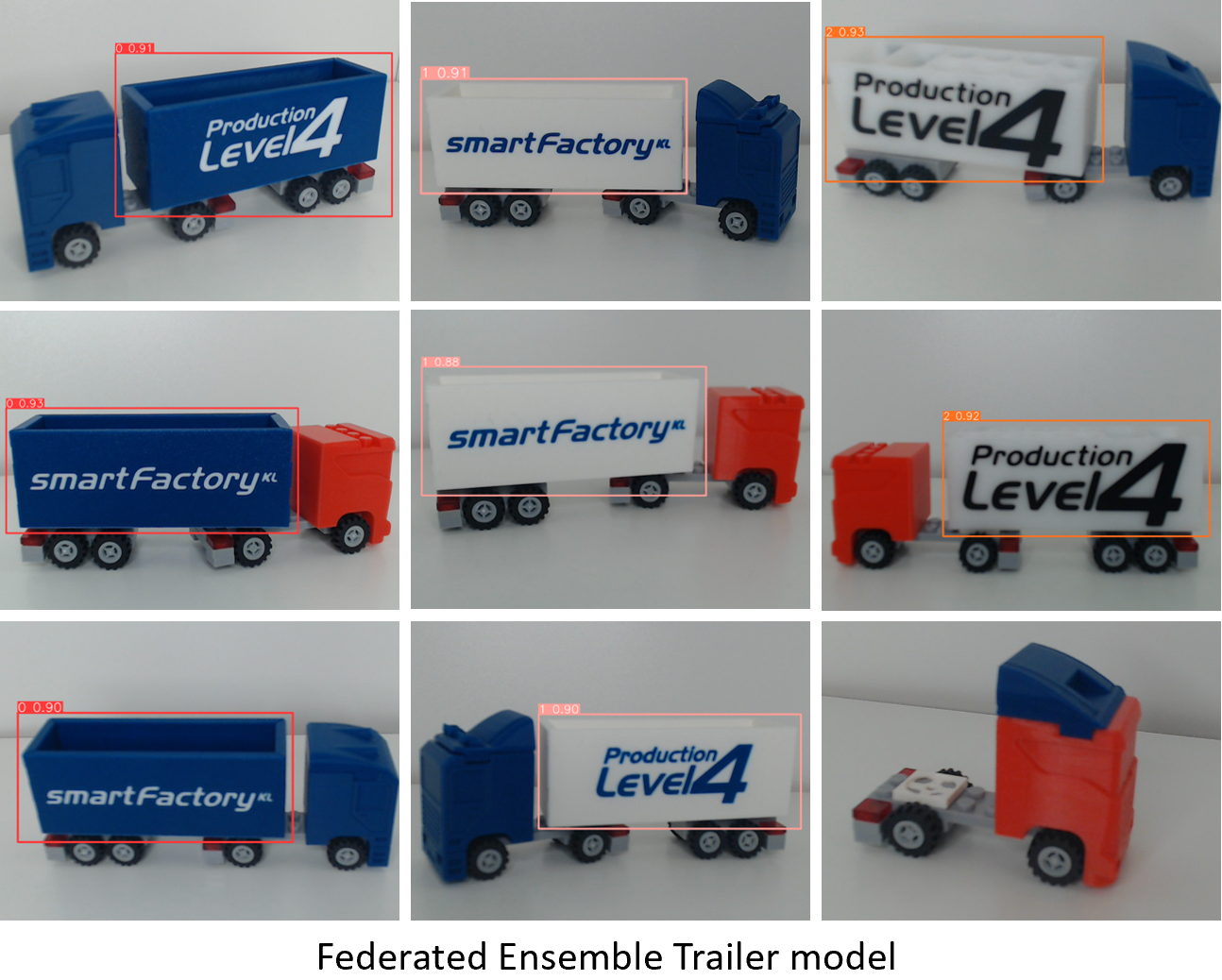}
    \caption{Output of trailer model trained with Federated Ensemble YOLOv5 algorithm on cabins assembled with different trailer types (0: Trailer\_blue, 1: Trailer\_white, 2: Trailer\_white\_penholder)}
    \label{fig:experiment5_2}
\end{figure}

The test images used in Experiment 5 correspond to the same set as those employed in Experiment 2. Figure \ref{fig:experiment5} showcases the output of the YOLOv5 trailer model on these images, revealing its poor performance. For instances featuring a blue trailer with a blue cabin combination, the model erroneously predicts the entire truck as 'trailer\_body\_blue'. Furthermore, the predicted trailer bounding box encompasses the chassis and exhibits a significant number of false positives across various images. In stark contrast, the results obtained from the FedEnsemble trailer model are truly remarkable. The FedEnsemble model avoids generating any false positives and accurately predicts and draws precise bounding boxes exclusively around the trailer object, effectively excluding the chassis from the prediction, as shown in Figure \ref{fig:experiment5_2}.

In the final experiment, the trailer models are subjected to testing using images from the demonstrator. Similar to Experiment 3, these test images exhibit distinct backgrounds and lighting conditions compared to the trailer's training dataset. Once again, the performance of the YOLOv5 trailer model on these images is unsatisfactory. The model fails to predict the objects accurately, whereas only identifying the white trailer, that too with an imprecise bounding box. Additionally, the model produces a false positive on an image without any objects present.
In contrast, the FedEnsemble YOLOv5 trailer model excels in this challenging scenario. It not only correctly classifies the objects but also draws precise bounding boxes around them, even in an unfamiliar environment characterized by different lighting conditions that may affect the color appearance of the objects. The FedEnsemble model's ability to generalize well and maintain accurate predictions in such conditions demonstrates its superior performance compared to the standard YOLOv5 model.

\begin{figure}
    \centering
    \includegraphics[width= 0.48\textwidth]{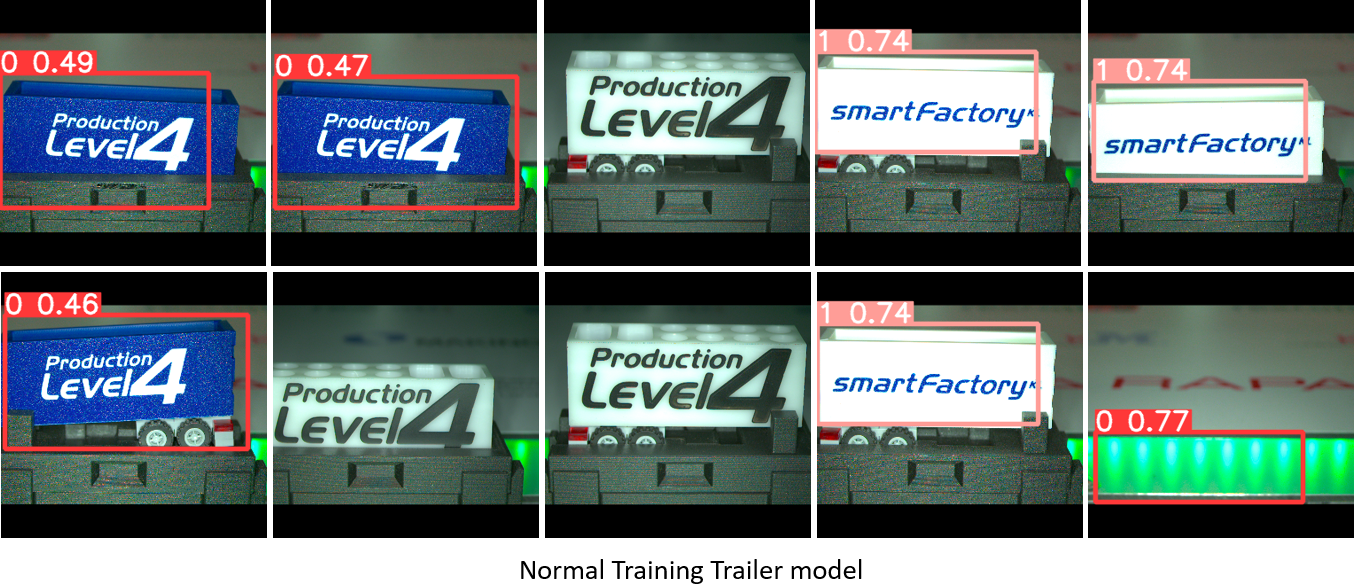}
    \caption{Output of model trained with Normal YOLOv5 algorithm on images from the Demonstrator (0: Trailer\_blue, 1: Trailer\_white, 2: Trailer\_white\_penholder)}
    \label{fig:experiment6}
\end{figure}

\begin{figure}
    \centering
    \includegraphics[width= 0.48\textwidth]{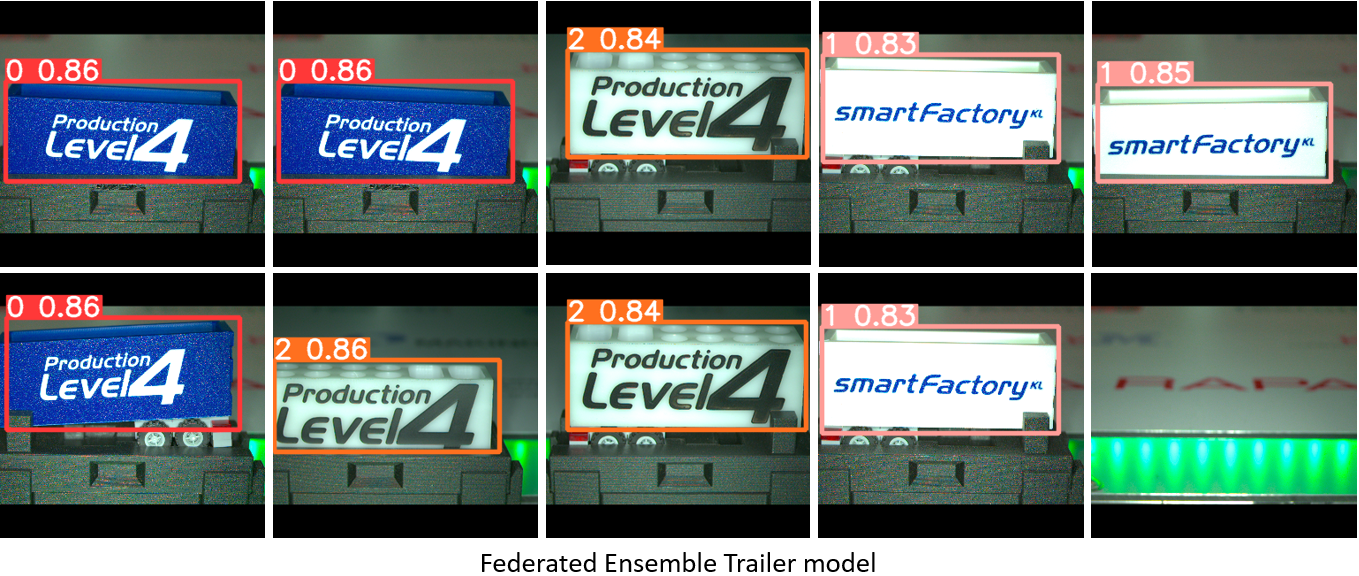}
    \caption{Output of model trained with Federated Ensemble YOLOv5 algorithm on images from the Demonstrator (0: Trailer\_blue, 1: Trailer\_white, 2: Trailer\_white\_penholder)}
    \label{fig:experiment6_2}
\end{figure}

\section{Discussion}
In this section, we discuss the implications and significance of the results obtained from the experiments conducted in this study.
The experimental results clearly demonstrate the superiority of the federated ensemble approach over the centralized training approach in the context of object detection using the YOLOv5 algorithm. The federated ensemble YOLOv5 model consistently outperformed the centralized YOLOv5 model across various scenarios for both cabin and trailer datasets.
One key advantage of the federated ensemble approach is its ability to effectively leverage the collective knowledge of multiple decentralized models. By aggregating the predictions from individual models trained on local data, the federated ensemble model achieves higher accuracy, better bounding box predictions, and reduced false positives compared to the centralized trained model.

The experiments involving different combinations of cabins and windshields, as well as trailers with and without chassis, showcased the robustness of the federated ensemble model. It successfully classified and accurately predicted objects, even when presented with combinations not present in the training dataset. This indicates the model's ability to generalize well and handle variations in object appearance and orientation.
Moreover, the tests conducted on images from the demonstrator, which featured different background settings and lighting conditions, further demonstrated the effectiveness of the federated ensemble approach. The model consistently achieved accurate classifications and precise bounding box predictions, even in challenging and unfamiliar environments. This highlights the Federated ensemble algorithm's adaptability and potential for real-world applications.

\subsection{Potential for Generalization and Future Research}

The positive outcomes obtained from this research indicate the potential applicability of the federated ensemble approach beyond the YOLOv5 algorithm and object detection tasks. The concept of federated learning, combining the strengths of ensemble methods and preserving data privacy, can be extended to other architectural models and various use cases, including image classification, image segmentation, and small object detection.
Furthermore, the potential of applying federated learning to non-vision tasks, such as spam detection and anomaly detection, warrants exploration. By decentralizing the training process and aggregating models' predictions, federated learning may provide improved performance and data privacy in these domains as well.
Validation of these results through future experiments on public datasets would further support the effectiveness of the federated ensemble approach and solidify its potential for practical implementation.

\section{Conclusion}
Following the realization, that federated learning acts as a combination of Bagging and Boosting algorithms.
This was empirically tested using a YOLOv5 algorithm, specifically tailored to our research context. The comparative results highlighted a superior performance of the federated ensemble YOLOv5 algorithm, in contrast to the centralized YOLOv5 model, particularly within the context of a custom dataset applied to manufacturing scenarios.
This approach is not limited to YOLOv5, but can be applied to other object detection algorithms as well. Future research will explore the potential of applying this federated ensemble approach to diverse architectural models and use-cases, including but not limited to, image classification, image segmentation, and small object detection. Moreover, we also aim to extend its applicability to non-vision use cases, such as spam detection, anomaly detection, and more. There is potential for the validation of these results through future application on public datasets, which would further substantiate our findings.
In conclusion, our research demonstrates that, in addition to the well-known advantages of preserving data sovereignty, FL also holds potential advantages in situations where access to a substantial portion of the data set is possible.

\bibliographystyle{IEEEtran.bst}
\bibliography{lit}


\end{document}